\documentclass[runningheads]{llncs}
\usepackage[T1]{fontenc}
\usepackage{graphicx}
\usepackage{amsfonts}
\usepackage{amsmath}
\usepackage{orcidlink}
\usepackage{cleveref}
\usepackage[sort]{cite}
\usepackage{siunitx}
\usepackage{setspace}
\usepackage{nicefrac}
\usepackage{tikz}
\usetikzlibrary{decorations.text}
\usetikzlibrary{decorations.markings}
\usetikzlibrary{calc}
\usetikzlibrary{shapes}
\usetikzlibrary{positioning}
\usetikzlibrary{arrows}
\usetikzlibrary{shapes.arrows}
\usetikzlibrary{arrows.meta}
\usetikzlibrary{fit}
\usetikzlibrary{shapes.geometric}
\usetikzlibrary{backgrounds}

\newlength{\minimumsizeaicon}
\newlength{\minimumheightaicon}
\newlength{\minimumsizeinterconnection}
\newlength{\minimumheightinterconnection}

\definecolor{aicRed}{HTML}{e78284}
\definecolor{aicGreen}{HTML}{a6d189}
\definecolor{aicBlue}{HTML}{8caaee}
\definecolor{aicOrange}{HTML}{e5c890}
\definecolor{aicGray}{HTML}{dce0e8}

\tikzset{
	aicon/state/.style={
		rectangle, fill=aicBlue, rounded corners=4pt,
		minimum width=\minimumsizeaicon, minimum height=\minimumheightaicon,
		inner sep=0pt, text=white, align=center,
		font=\bfseries\linespread{0.9}\selectfont
	},
	aicon/active/.style={
		rectangle, rounded corners=4pt, fill=aicGreen,
		minimum width=\minimumsizeinterconnection,
		minimum height=\minimumheightinterconnection,
		inner sep=0pt, text=white, align=center,
		font=\linespread{0.9}\selectfont
	},
	aicon/sensor/.style={
		rectangle, rounded corners=4pt, fill=aicOrange,
		minimum width=\minimumsizeaicon, minimum height=\minimumheightaicon,
		inner sep=0pt, text=white, align=center, font=\bfseries
	},
	aicon/goal/.style={
		diamond, fill=aicRed, minimum size=20pt,
		inner sep=0pt, text=white, font=\bfseries\footnotesize
	},
	aicon/link/.style={draw=aicGray, line width=2.5pt},
	aicon/gradient/.style={draw=aicRed, line width=2.5pt, ->, >=latex}
}

\begin{document}
	
	\title{A Mechanistic Model for Collective Motion from Sensorimotor Regularities}
	
	\titlerunning{A Mechanistic Model for Collective Motion from Sensorimotor Regularities}
	
	\author{Vito Mengers\inst{1,2}\orcidlink{0000-0001-8341-0299} \and
		Bao Duc Cao\inst{1} \and
		Oliver Brock\inst{1,2,3}\orcidlink{0000-0002-3719-7754}}
	
	\authorrunning{V. Mengers et al.}
	
	\institute{Robotics and Biology Laboratory, Technische Universit\"{a}t
		Berlin, Germany \and
		Science of Intelligence, Research Cluster of Excellence, Berlin,
		Germany \and
		Robotics Institute Germany \\
		\url{https://www.tu.berlin/robotics/papers/collective-motion}}
	
	\maketitle
	
	\begin{abstract}
		Collective behavior in animals has long been modeled through
		self-propelled particle models, which reproduce striking group-level
		phenomena through abstract interaction forces. Yet these models are
		fundamentally descriptive: they leave open the question of how
		collective behavior is actually produced. Recent empirical work makes
		this gap concrete: locusts do not align with neighbors, sensory and
		cognitive mechanisms mediate interaction instead. A mechanistic model
		must therefore operate at the sensorimotor level, grounded in what
		individual organisms can actually perceive, estimate, and physically
		execute. We present such a model based on a modeling framework from
		robotics, extended here to collective motion. Each agent perceives
		neighbors through bearing and apparent-size cues within a limited field
		of view, maintains uncertain internal state estimates, and selects
		actions through gradient descent on a desired social
		distance---without any prescribed interaction forces. This simple model
		produces diverse collective behaviors including polarized motion,
		milling, ring formations, and subgroup fragmentation. A global
		sensitivity analysis shows that behavioral transitions are governed by
		sensorimotor parameters corresponding to measurable biological
		quantities: field of view geometry, sensory noise, turning agility, and
		memory. Collective behavior can therefore be understood as the emergent
		outcome of interacting sensorimotor regularities, and differences
		across species as the emergent outcome of differences in embodiment and
		environment.
		
	\end{abstract}
	
	\section{Introduction}
	Collective behavior is one of the most striking phenomena in the natural world. Flocks of birds~\cite{ballerini2008interaction,cavagna2010scale}, schools of fish~\cite{katz2011inferring,rosenthal2015revealing}, and swarms of insects~\cite{buhl2006disorder,attanasi2014collective} produce coherent group-level patterns from purely local interactions, without centralized control or global information. The mystery is not just that groups coordinate, but that they do so with so little: each individual sees only a handful of neighbors, yet the group behaves as if guided by something more. It is this question---how so much order arises from so little---that has driven a rich tradition of computational modeling in biology, physics, and robotics~\cite{vicsek2012collective,sumpter2010collective}.
	
	Self-propelled particle models~\cite{vicsek1995novel,couzin2005effective} have been the dominant modeling framework for collective behavior. Agents align with neighbors within a fixed radius, subject to noise, and the result is polarization, milling, cohesion, and phase transitions~\cite{ginelli2016physics}---establishing collective behavior as a field of rigorous quantitative inquiry with broad influence~\cite{marchetti2013hydrodynamics}. These models are compelling because of their simplicity, but that simplicity comes at a cost. Their parameters---interaction radii, alignment strengths, noise levels---are fit to observed patterns and carry no necessary correspondence to biological quantities. They characterize what collective behavior looks like rather than how it is produced and assume agents have direct access to neighbors' positions and headings, quantities that may not be available under realistic sensory constraints~\cite{strandburg2013visual}. Recent empirical work makes this concrete: Sayin et al.~\cite{sayin2025behavioral} showed that locusts do not align with neighbors at all; sensory and cognitive mechanisms mediate interaction instead. This calls for a different level of modeling: one grounded in what organisms can actually perceive and physically execute, with parameters corresponding to measurable biological quantities, so that predictions transfer to new species without refitting.
	
	We present such a mechanistic model for collective behavior based on \textit{Active InterCONnect} (AICON)~\cite{mengers2025noplan,martinmartin2022coupled}, a framework from robotics that generates behavior by composing sensorimotor regularities without directly encoding behavior itself. This makes it a natural substrate for a mechanistic model of collective behavior. The components that determine how an agent perceives, estimates, and acts are precisely the quantities that vary across species and environments. AICON has been applied to biological information processing~\cite{mengers2025scanpath}, including collective opinion dynamics~\cite{mengers2024leveraging}. Here we extend it to collective motion. Each agent perceives neighbors through bearing and apparent-size cues within a limited field of view, maintains uncertain internal state estimates, and selects actions through gradient descent on a desired social distance---without any prescribed interaction forces. Unlike models grounded in instantaneous visual responses~\cite{bastien2020model,mezey2024visual}, abstract Bayesian objectives~\cite{heins2024collective}, or specific neural implementations~\cite{salahshour2025allocentric}, AICON maintains persistent uncertain representations and operates at the sensorimotor level without prescribing how perception translates into social response.
	
	The model recovers the behavioral diversity that particle models reproduce---polarized motion, milling, ring formations, and subgroup fragmentation---but from sensorimotor parameters alone: field of view, sensory noise, and turning agility. These are not fitted constants but measurable biological quantities. The same parameters that govern coordination also govern its breakdown, and varying them generates predictions about how collective behavior should differ across species with different sensory systems, motor capabilities, and environments.

	\section{A Sensorimotor Model of Collective Motion}
	
	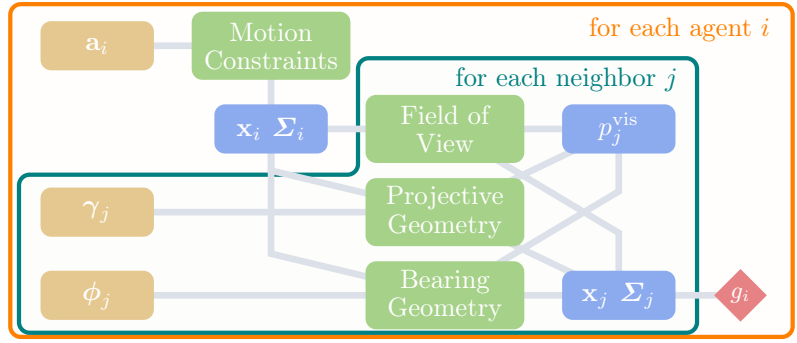
\begin{figure}[t]
		\centering
		\small
		\newlength{\cnodedisty}
		\newlength{\cnodedistx}
		\setlength{\cnodedisty}{0.9cm}
		\setlength{\cnodedistx}{2.3cm}
		\pgfdeclarelayer{bg}
		\pgfsetlayers{bg,main}
		\begin{tikzpicture}[on grid,
			node distance=\cnodedisty and \cnodedistx]
			
			\node[aicon/sensor] (zphi)   {$\boldsymbol{\phi}_j$};
			\node[aicon/sensor, above=of zphi]  (zgamma) {$\boldsymbol{\gamma}_j$};
			\node[aicon/sensor, above=2\cnodedisty of zgamma] (aego) {$\mathbf{a}_{i}$};
			
			\node[aicon/active, right=of aego] (motconstr) {Motion\\ Constraints};
			\node[aicon/state, below=of motconstr] (xi) {$\mathbf{x}_{i}$ $\boldsymbol{\Sigma}_{i}$};
			
			\node[aicon/active, right=of xi] (fov) {Field of\\ View};
			\node[aicon/active, below=of fov] (projgeom) {Projective\\ Geometry};
			\node[aicon/active, below=of projgeom] (bearinggeom) {Bearing\\ Geometry};
			
			\node[aicon/state, right=of bearinggeom] (xj) {$\mathbf{x}_{j}$ $\boldsymbol{\Sigma}_{j}$};
			\node[aicon/state, above=2\cnodedisty of xj] (pvis) {$p^{\mathrm{vis}}_{j}$};
			
			\node[rectangle, above=0.6\cnodedisty of fov, align=left,
			text=teal, xshift=0.7\cnodedistx] (label1) {for each neighbor $j$};
			\node[rectangle, above=0.6\cnodedisty of label1, align=right,
			text=orange, xshift=0.65\cnodedistx] (label2) {for each agent $i$};
			
			\node[aicon/goal, right=0.7\cnodedistx of xj] (g) {$g_i$};
			
			\begin{pgfonlayer}{bg}
				\draw[rounded corners, draw=orange, ultra thick, fill=orange, fill opacity=0.01]
				([xshift=-0.5\cnodedistx,yshift=-0.5\cnodedisty]zphi.center)
				-- ([xshift=1.0\cnodedistx, yshift=-0.5\cnodedisty]xj.center)
				-- ([xshift=1.0\cnodedistx, yshift=1.5\cnodedisty]pvis.center)
				-- ([xshift=-0.5\cnodedistx, yshift=0.5\cnodedisty]aego.center)
				-- cycle;
				
				\draw[rounded corners, draw=teal, ultra thick, fill=teal, fill opacity=0.01]
				([xshift=-0.45\cnodedistx,yshift=-0.45\cnodedisty]zphi.center)
				-- ([xshift=0.45\cnodedistx, yshift=-0.45\cnodedisty]xj.center)
				-- ([xshift=0.45\cnodedistx, yshift=0.85\cnodedisty]pvis.center)
				-- ([xshift=-0.5\cnodedistx, yshift=0.85\cnodedisty]fov.center)
				-- ([xshift=-0.5\cnodedistx, yshift=-0.55\cnodedisty]fov.center)
				-- ([xshift=-0.45\cnodedistx, yshift=0.45\cnodedisty]zgamma.center)
				-- cycle;
				
				\draw[aicon/link] (aego.center) -- (motconstr.center);
				\draw[aicon/link] (motconstr.center) -- (xi.center);
				\draw[aicon/link] (zphi.center) -- (bearinggeom.center);
				\draw[aicon/link] (zgamma.center) -- (projgeom.center);
				\draw[aicon/link] (xi.center) -- ([yshift=-1.5\cnodedisty]xi.center)
				-- (bearinggeom.center);
				\draw[aicon/link] (xi.center) -- ([yshift=-0.5\cnodedisty]xi.center)
				-- (projgeom.center);
				\draw[aicon/link] (xi.center) -- (fov.center);
				\draw[aicon/link] (pvis.center) -- ([yshift=-0.75\cnodedisty]pvis.center)
				-- (bearinggeom.center);
				\draw[aicon/link] (pvis.center) -- (projgeom.center);
				\draw[aicon/link] (bearinggeom.center) -- (xj.center);
				\draw[aicon/link] (projgeom.center) -- (xj.center);
				\draw[aicon/link] (xj.center) -- ([yshift=0.75\cnodedisty]xj.center)
				-- (fov.center);
				\draw[aicon/link] (fov.center) -- (pvis.center);
				\draw[aicon/link] (xj.center) -- (g.center);
			\end{pgfonlayer}
		\end{tikzpicture}
		\caption{Collective behavior emerges from the composition of local
			sensorimotor regularities, with no interaction forces prescribed at
			the collective level. Each agent~$i$ maintains recursive estimators
			for its own pose $(\mathbf{x}_i, \boldsymbol{\Sigma}_i)$ and, for
			each neighbor~$j$, for position $(\mathbf{x}_j,\boldsymbol{\Sigma}_j)$
			and visibility $p^{\mathrm{vis}}_j$. Active interconnections couple
			these states: \textit{Bearing Geometry} and \textit{Projective
				Geometry} update neighbor position from observations $\phi_j$ and
			$\gamma_j$, modulated by visibility and own pose; \textit{Field of
				View} computes visibility from own and neighbor states;
			\textit{Motion Constraints} propagate pose from
			actions~$\mathbf{a}_i$. The goal~$g_i$, minimizing expected deviation from a desired social distance $d_0$, drives action selection through gradient descent along three emergent paths: position adjustment toward $d_0$, active sensing to reduce estimation uncertainty, and reorientation to maintain neighbors within the field of view---none of which are explicitly encoded.}
		\label{fig:aicon_collective}
	\end{figure}
	
	AICON~\cite{mengers2025noplan} generates behavior through the dynamic composition of sensorimotor regularities, without directly encoding behavior itself. It provides two structural elements: \textit{recursive estimators} that maintain probabilistic beliefs about task-relevant quantities over time, and \textit{active interconnections} that encode regularities in the relationships between those quantities as functional dependencies. By composing these elements, AICON builds a network through which gradients can be propagated from goals back to actions. Behavior emerges from following these gradients---not from a predefined policy or any training, but from the structure of the sensorimotor regularities and the current state of the world---so the same architecture operates across all parameter settings without modification.
	
	We instantiate this architecture for collective motion as shown in \Cref{fig:aicon_collective}. Each agent maintains estimators for its own pose and, for each neighbor, for position and visibility. Neighbors are perceived through two angular cues: bearing $\phi$, the direction to the neighbor relative to the agent's heading, and apparent size $\gamma$, the angular size of the neighbor as seen by the agent. Together these encode directional and distance-related information without requiring explicit depth sensing, grounding interaction in what the agent can actually observe. Perception is constrained by a limited field of view $\psi$: the visibility state $p^\mathrm{vis}$ is computed from ego and neighbor position estimates and modulates how strongly observations update the neighbor belief. When a neighbor leaves the field of view, its estimator continues propagating the last belief forward, accumulating uncertainty. The flow of social information is therefore explicitly state- and uncertainty-dependent---a structural consequence of composing these regularities, not a prescribed rule.
	
	Behavior arises from a single cost term: each agent $i$ minimizes the expected deviation
	from a desired social distance $d_0$ across all estimated
	neighbors~$\mathcal{V}_i$,
	\begin{equation}
		g_i = \sum_{j \in \mathcal{V}_i}
		\mathbb{E}\!\Big[\big|\,\|\mathbf{x}_j\| - d_0\big|\Big],
	\end{equation}
	
	\noindent
	where the expectation over the belief implicitly couples the cost to estimation uncertainty over the neighbor's relative position $\mathbf{x}_j$. This is the only objective, yet when pursued with simple gradient descent through the model it gives rise to three distinct gradient paths without any of them being explicitly programmed. The first adjusts position to reduce deviation from $d_0$. The second flows through uncertainty: since uncertainty enters the expected cost, there is a gradient to reduce it---producing active sensing and triangulating motions without this being encoded anywhere. The third flows further still: because the uncertainty is modulated by $p^\mathrm{vis}$, there is a gradient to reorient toward neighbors near the field of view boundary. This gradient ensures visibility and thus group cohesion even under narrow fields of view, without implicit encoding. Of these gradients for different neighbors each agent selects the steepest at each time step~\cite{mengers2025noplan}, while angular velocity is bounded  by~$\omega_\mathrm{max}$, enforcing a physical turning constraint that,
	as we will show, has a dominant influence on emergent collective
	structure.
	
	No interaction forces are prescribed at the collective level. The desired-distance objective induces an effective coupling between agents, but this emerges from the sensorimotor architecture rather than being specified as a pairwise force law. Coordination emerges entirely from agents following local gradients through their own sensorimotor regularities, interacting through shared space and mutual perception, yielding a simple mechanistic model that, as we show, produces a diverse range of collective behaviors governed by sensorimotor parameters.
	
	\section{Experimental Setup}
	
	We simulate groups of $N = 250$ agents in an unbounded two-dimensional environment. Each agent is modeled as a circular body of unit radius with unicycle kinematics, controlled by linear velocity~$v$ and angular velocity~$\omega$, with $|\omega| \leq \omega_{\mathrm{max}}$. All simulations use a fixed timestep of $\Delta t = 0.1$ and a fixed desired social distance $d_0 = 50$. We vary sensing, estimation, and motor parameters as summarized in \Cref{tab:params}, running each configuration both with memory (neighbor estimates propagated during invisibility) and without (estimates discarded and re-initialized on re-entry). For all reported qualitative results and figures, we repeated the experiment across 10 starting configurations and random seeds. The reported regimes are robust across all these conditions with variation limited to the geometric tightness of compact formations.
	
	\begin{table}[t]
	\centering
	\caption{Varied parameters and their ranges.}
	\small
	\begin{tabular}{p{1.7cm}p{6cm}p{4cm}}
		\hline
		Parameter & Description & Values \\\hline \\[-7pt]
		$\psi$ & Field of view &
		$\bigl\{\tfrac{\pi}{2},\,\pi,\,\tfrac{3\pi}{2},\,2\pi\bigr\}$ \\[2pt]
		$\omega_{\mathrm{max}}$ & Maximum angular velocity &
		$\{0.1,\;0.3,\;0.6,\;0.9\}$ \\[2pt]
		$\sigma_{\phi}$ & Assumed bearing noise std.\ &
		$\{0.001,\;0.01,\;1,\;10,\;100\}$ \\[2pt]
		$\sigma_{\gamma}$ & Assumed apparent size noise std.\ &
		$\{0.001,\;0.01,\;1,\;10,\;100\}$ \\[2pt]
		$Q$ & Assumed process noise for other agents &
		$\{0.01,\;0.1,\;1,\;10\}$
	\end{tabular}
	\label{tab:params}
\end{table}

	\textbf{Metrics and Quantitative Analysis}\quad We evaluate emergent collective behavior using five complementary metrics that together capture the behavioral diversity the model aims to produce. \textit{Polarization}~$P$~\cite{vicsek1995novel} measures
	directional alignment as the magnitude of the mean unit velocity vector
	across all agents, averaged over time; values near~$1$ indicate
	coherent directed motion, while values near~$0$ indicate dispersed or
	rotational motion. \textit{Relative Circular Area}
	(RCA)~\cite{mezey2024visual} quantifies spatial compactness by
	measuring how closely the convex hull of agent positions resembles a
	circle; high RCA indicates compact circular formations, low RCA
	elongated or fragmented ones. \textit{Maximum mean
		displacement}~$D$ captures global translation as
	the maximum displacement of the group center of mass over the
	simulation. \textit{Center Distance}~$C$
	measures spatial cohesion as the time-averaged minimum distance from
	any agent to the group center of mass. \textit{Directional
		fragmentation} is quantified by applying
	DBSCAN~\cite{schubert2017dbscan} to agent headings in circular space,
	yielding the average number of heading clusters~$K$ and
	clustered fraction~$F$, capturing whether the
	population persistently splits into subgroups moving in distinct
	directions. To quantify parameter influence we compute Sobol
	first-order ($S_1$) and total-order ($S_T$) sensitivity
	indices~\cite{sobol2001global}; the gap between them indicates the
	degree to which a parameter acts through interactions rather than in
	isolation.
	
	\section{Behavioral Diversity from Sensorimotor Parameters}
	
		\begin{figure}[t]
		\centering
		\includegraphics[width=\textwidth]{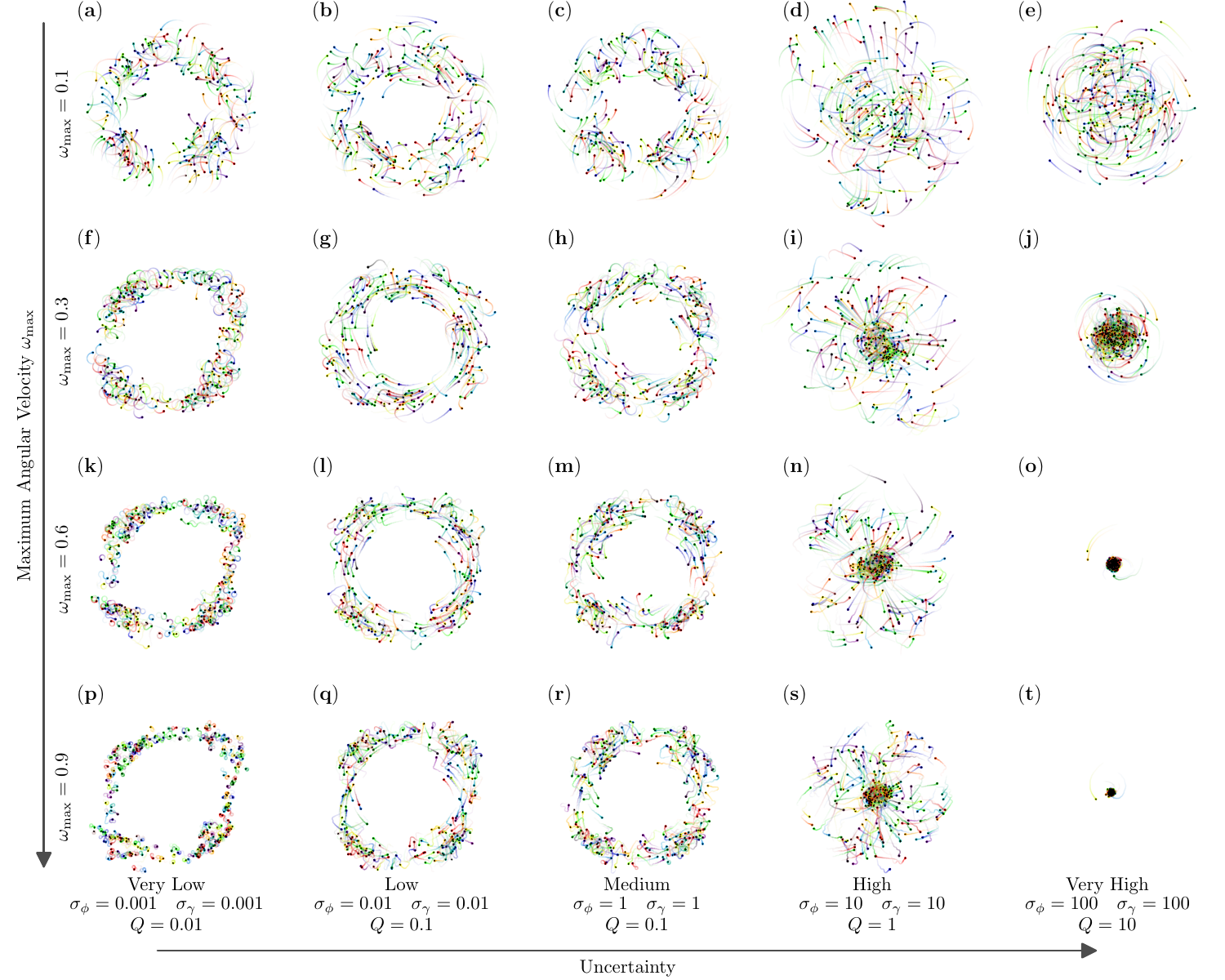}
		\caption{Turning agility and sensory uncertainty jointly determine
			collective structure, with high uncertainty collapsing all formations
			into dense clusters regardless of motor capability. At low-to-medium
			uncertainty (cols.\ 1--3), increasing $\omega_\mathrm{max}$ produces
			a graded transition from large smooth rings to tighter compact orbits.
			This breaks down at higher uncertainty (cols.\ 4--5): estimation
			error overwhelms the gradient signal sustaining rotational motion,
			collapsing all configurations into dense milling clusters whose
			tightness increases with $\omega_\mathrm{max}$.}
		\label{fig:wmax}
	\end{figure}
	\begin{figure}[t]
		\centering
		\includegraphics[width=\textwidth]{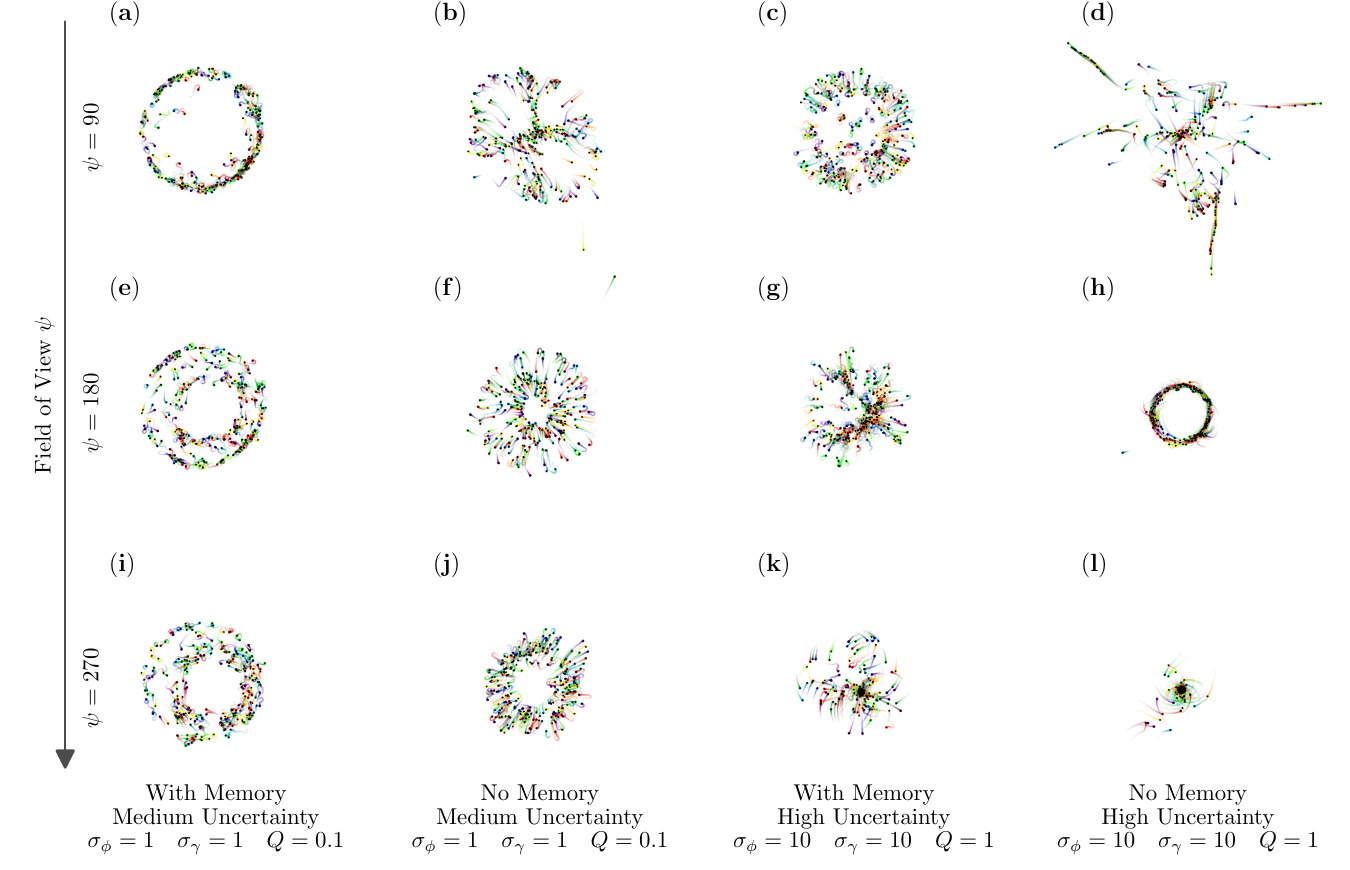}
		\caption{Memory is critical for cohesion under narrow fields of
			view, while its absence can produce qualitatively different but
			stable structures under wider fields. At $\psi = 90°$, memory
			yields a ring (\textbf{a}, \textbf{c}) while its absence causes
			complete fragmentation (\textbf{b}, \textbf{d}). As field of view
			widens, cohesion is preserved even without memory, though structures
			differ---compare the loose ring in (\textbf{e}) with the tight
			milling ring in (\textbf{h}).}
		\label{fig:fov}
	\end{figure}
	
	Starting from identical initial conditions, variation of sensorimotor parameters alone produces qualitatively distinct collective behaviors without any change to the underlying interaction architecture---exactly what a mechanistic model grounded in sensorimotor regularities should produce. \textit{Stable rings} emerge at low uncertainty, where agents maintain reliable neighbor estimates and accurate distance control; high turning agility keeps agents on circular orbits at~$d_0$ (\Cref{fig:wmax},~\textbf{p}), while lower agility produces the same pattern with larger, smoother orbits (\Cref{fig:wmax},~\textbf{a}). \textit{Milling} emerges under increased uncertainty, where degraded distance estimation prevents stable orbit maintenance; moderate uncertainty produces loose irregular rings (\Cref{fig:wmax},~\textbf{m}) while high uncertainty collapses agents into dense rotating clusters (\Cref{fig:wmax},~\textbf{j}). 
	
	\textit{Fragmentation into groups moving along a line} occurs under narrow field of view with high noise and no memory: without predictive belief propagation, agents preferentially follow whoever is directly ahead, producing elongated drifting groups (\Cref{fig:fov},~\textbf{d}). \textit{Fragmentation into parallel streams} occurs under similar conditions but lower noise: reliable estimates allow lateral spacing maintenance, but absent memory causes the group to split into streams (\Cref{fig:fov},~\textbf{b}). The same model, cost function, and architecture---only the sensorimotor parameters change, and each regime emerges without prescribing rules for it.
	
	Crucially, the same collective behavior also emerges when agents receive noiseless ground-truth measurements from the simulator while still assuming nonzero noise in their internal model. The qualitative regime is preserved; compact formations become slightly tighter as estimation converges more rapidly. This confirms that the collective behaviors are a consequence of the sensorimotor estimation architecture rather than of stochastic perturbations---in contrast to particle models, where noise is necessary for nontrivial collective dynamics.
	
	\begin{figure}[t]
		\centering
		\includegraphics[width=\textwidth]{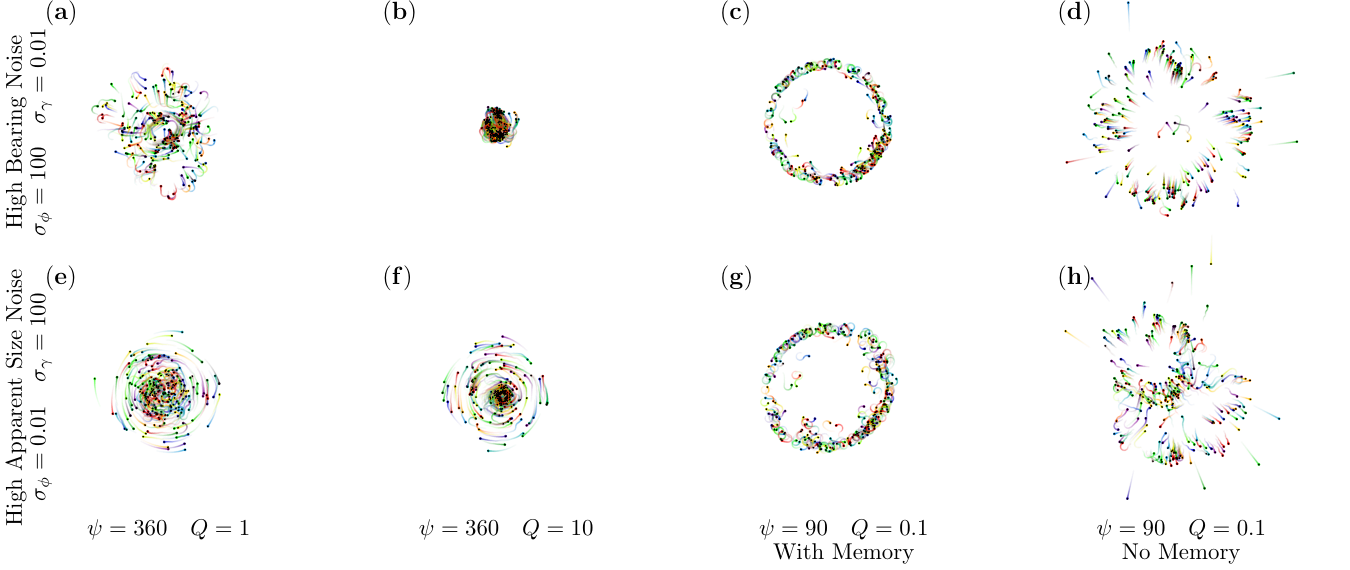}
		\caption{Bearing and apparent size noise degrade collective
			structure differently, reflecting distinct emergent compensation
			strategies. High bearing noise (top row) disrupts directional
			estimation, producing disordered clusters (\textbf{a}--\textbf{b}),
			though low process noise enables recovery through temporal
			integration and active triangulation (\textbf{a}, \textbf{c}). High
			apparent size noise (bottom row) leaves directional information
			intact but degrades distance estimation, producing inward-spiraling
			rings via distance-reducing triangulation (\textbf{e}--\textbf{f}).
			Without memory, both regimes lose coherence and fragment into
			aligned subgroups (\textbf{d}, \textbf{h}). Bearing and apparent
			size noise are thus not interchangeable---a distinction no scalar
			noise model could capture.}
		\label{fig:channel}
	\end{figure}
	
	\subsection{Sensorimotor Parameters Govern Behavioral Transitions}
	
	We now examine how three biologically interpretable parameter groups
	shape the behavioral regimes described above.
	
	\textbf{Motor constraints}\quad$\omega_\mathrm{max}$ governs formation geometry when estimation is reliable but becomes less relevant at high uncertainty. Low angular velocity produces smooth large-radius rings; higher values produce tighter orbits (\Cref{fig:wmax}). Beyond a threshold uncertainty, $\omega_\mathrm{max}$ instead determines only the tightness of the resulting milling cluster (\Cref{fig:wmax}, last two rows). This parameter has no counterpart in classical particle models, where heading changes are instantaneous and unconstrained. Crucially, it is not a fitted constant but a measurable property of the organism---the kind of biologically grounded parameter a mechanistic model should be governed by.
	
	\textbf{Perceptual uncertainty}\quad Bearing and apparent size noise degrade collective structure through qualitatively different mechanisms. High bearing noise disrupts directional localization, driving a transition toward disordered aggregation (\Cref{fig:channel},~\textbf{a}--\textbf{b}). High apparent size noise instead degrades distance estimation while leaving directional coordination intact, producing inward-spiraling formations as agents triangulate to recover distance information (\Cref{fig:channel},~\textbf{e}--\textbf{f}). Where process noise is low, agents partially compensate through temporal integration and active sensing. That bearing and apparent size noise produce distinct behavioral signatures reflects the distinct roles of these two sensory channels---a distinction invisible to models that treat noise as a scalar perturbation.
	
	\textbf{Memory}\quad Memory determines whether gradient continuity is preserved across periods of visual occlusion, and its effect is qualitative rather than graded. Under narrow fields of view, removing memory causes complete fragmentation despite identical parameters and initial conditions (\Cref{fig:fov}, first row). Under wider fields, memory loss shifts the attractor toward qualitatively different stable formations (\Cref{fig:fov},~\textbf{b}). Whether an organism retains estimates of occluded neighbors is a property of its cognitive architecture---and here it determines qualitatively different collective regimes, not just quantitative variation.
	
	\subsection{Global Sensitivity of Collective Behavior}
	
\begin{figure}[t]
		\centering
		\includegraphics[width=\textwidth]{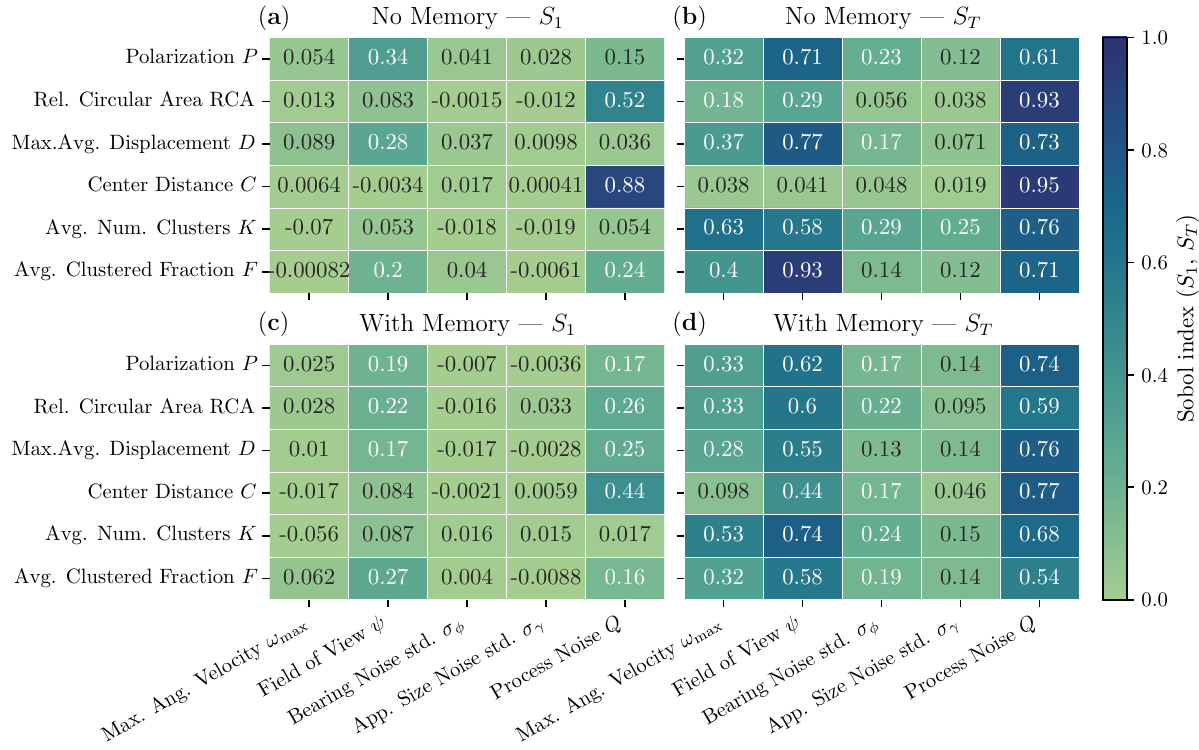}
		\caption{Field of view and overall uncertainty are the dominant
			direct drivers of collective behavior, while memory redistributes
			sensitivity away from individual parameters. Only $\psi$ and $Q$
			show substantial first-order contributions ($S_1$); $\sigma_\phi$,
			$\sigma_\gamma$, and $\omega_\mathrm{max}$ instead act primarily
			through interactions (low $S_1$, elevated $S_T$). With memory
			(\textbf{c}--\textbf{d}), sensitivity is more evenly distributed,
			suggesting predictive belief propagation absorbs variance that would
			otherwise concentrate on $\psi$ and $Q$.}
		\label{fig:sobol}
	\end{figure}

	The qualitative patterns above are confirmed and quantified by a global variance-based sensitivity analysis (\Cref{fig:sobol}). Field of view $\psi$ and process noise $Q$ are the strongest direct drivers across nearly all metrics and both memory conditions, consistent with $\psi$'s role as the primary gating factor for neighbor visibility and gradient continuity, and with $Q$'s role in shaping spatial cohesion through overall estimation uncertainty. Sensory noise parameters $\sigma_\phi$ and $\sigma_\gamma$ show similarly modest first-order contributions but elevated total-order indices, indicating they act primarily through interactions. The angular velocity limit~$\omega_{\mathrm{max}}$ contributes consistently but modestly in isolation, in line with its role as a shaping rather than gating parameter. Memory substantially flattens the sensitivity landscape: first- and total-order indices become more evenly distributed, indicating that predictive belief propagation reduces the system's dependence on any single sensorimotor quantity and distributes sensitivity more evenly across the full parameter space. Taken together, these results confirm that behavioral transitions are governed by sensorimotor quantities---field of view, noise, turning agility, memory---rather than abstract tuning parameters, and that their influence operates through the coupled structure of perception, estimation, and action rather than in isolation. Sobol indices were estimated from a single sample; occasional negative first-order indices reflect estimation variance rather than true negative contributions and should be interpreted with caution.
	
	\section{Discussion}
	
	This work presents collective behavior as the emergent outcome of
	interacting sensorimotor loops, grounded in what individual agents can perceive, estimate, and physically execute. A diverse range of behaviors arises from variation of sensorimotor parameters alone. Behavioral transitions and coordination breakdowns are governed by parameters with direct biological interpretations, connecting cross-species variation and failure modes within a single explanatory~framework.
	
	\textbf{Comparison to Particle Models}\quad
	The core limitation of classical particle models is not that they are
	wrong about collective phenomena---\emph{they are not}---but that they
	are descriptive rather than mechanistic. Their parameters are free
	constants with no necessary correspondence to biological quantities,
	meaning they cannot transfer predictions to new species or environments
	without refitting, and cannot explain why coordination breaks down
	under specific sensory or motor conditions. As Sayin et al.~\cite{sayin2025behavioral} demonstrated, these mechanistic assumptions are not empirically supported even for a paradigmatic species. Our results demonstrate a
	constructive alternative: behavioral diversity is a prediction of the
	sensorimotor architecture, and the parameters governing transitions
	correspond to quantities measurable independently of collective behavior
	experiments.
	
	\textbf{Comparison to Perceptual Models}\quad
	Several recent models move beyond particle models by incorporating
	perceptual mechanisms. Vision-based response
	models~\cite{bastien2020model,mezey2024visual} ground sensory input in
	visual geometry, but their response parameters remain phenomenological
	and the models are stateless and memoryless---there is no estimation, uncertainty, or mechanism for behavior to depend on prior
	observations. Active inference models~\cite{heins2024collective} ground
	coordination in Bayesian belief updating, a genuine advance, but the
	generative model uses abstract sector-wise representations and
	parameters with no biological referent; motor constraints are
	absent and field of view is approximated rather than geometrically
	grounded. Ring attractor models~\cite{salahshour2025allocentric}
	connect collective behavior to neural architecture, but the empirical
	evidence for ring attractor dynamics applies specifically to heading
	direction encoding in \textit{Drosophila}~\cite{seelig2015neural}; whether social bearing to
	neighbors is encoded through the same architecture remains open for
	virtually all collective behavior species. Our model differs from all
	three: it operates at the sensorimotor level with parameters
	corresponding to measurable biological quantities, maintains persistent
	uncertain internal representations, and couples perception to action
	through sensorimotor geometry rather than abstract rules or neural
	commitments. The active sensing behavior that emerges---agents
	triangulating to reduce bearing uncertainty and reorienting to maintain
	neighbors within the field of view---cannot arise from memoryless
	response laws or abstract belief updating, and our results
	demonstrate that behavioral transitions are governed by biological quantities rather than fitted constants.
	
	\textbf{Limitations}\quad
	The cross-species prediction claim remains a hypothesis: parameter regimes corresponding to plausible biological values produce qualitatively distinct collective behaviors, but these predictions have not yet been validated against empirical data from specific organisms. In the current instantiation, neighbor interactions are also not geometrically localized: all estimated neighbors contribute to the cost regardless of occlusion or spatial configuration, which likely constrains observable structures toward ring-like formations. Empirical work suggests that visual neighborhoods based on ray-casting better account for individual responses than metric or topological alternatives~\cite{rosenthal2015revealing}, motivating geometrically localized interaction as a natural next extension alongside neighbor velocity estimation, richer agent dynamics incorporating acceleration constraints, validation against species-specific behavioral data, and structured environmental perturbations such as attraction points or external flow fields. The present model also treats all agents as identical; incorporating individual variation in sensorimotor parameters would bring it closer to biological reality and may reveal additional phenomena. The desired distance $d_0$ remained fixed throughout; we expect it to scale formation geometry without changing the qualitative regime structure, though this remains to be verified systematically.
	
	\section{Conclusion}
	
	We have shown that a diverse range of collective behaviors---polarized
	motion, milling, line formation, and subgroup formation---emerges from
	the composition of individual sensorimotor regularities, without
	prescribing interaction forces or fitting descriptive parameters to
	observed group patterns. Behavioral transitions are governed by
	parameters corresponding to measurable biological quantities---turning
	agility, field of view geometry, sensory noise, and memory---the same
	parameters that determine both successful coordination and its
	breakdown. This demonstrates that collective behavior is not a phenomenon
	requiring special interaction mechanisms, but a consequence of
	embodied agents pursuing simple goals under sensorimotor
	constraints, and that differences across species can productively be
	understood as differences in embodiment and environment.
	
	\begin{credits}
		\subsubsection{\ackname}
		We thank Pawel Romanczuk for valuable discussions and feedback on this work.
		We gratefully acknowledge funding by the Deutsche
		Forschungsgemeinschaft (DFG, German Research Foundation) under
		Germany's Excellence Strategy -- EXC~2002/1 ``Science of
		Intelligence'' -- project number 390523135. This work has been
		partially supported by the German Federal Ministry of Research,
		Technology and Space (BMFTR) under the Robotics Institute Germany
		(RIG).
		We used Claude Sonnet 4.6 (Anthropic) to suggest textual
		improvements, particularly to ensure conciseness.
		
		\subsubsection{\discintname}
		The authors declare no competing interests.
	\end{credits}
	
	\bibliographystyle{splncs04}
	\bibliography{aicon_collectives}
	
\end{document}